\documentclass[conference]{IEEEtran}
\IEEEoverridecommandlockouts
\usepackage{cite}
\usepackage{amsmath,amssymb,amsfonts}
\usepackage{algorithmic}
\usepackage{graphicx}
\usepackage{textcomp}
\usepackage{xcolor}
\usepackage{subcaption}

\usepackage{tabularx}
\usepackage{array}

\newcolumntype{Y}{>{\centering\arraybackslash}X}
\newcolumntype{Z}{>{\raggedright\arraybackslash}X}

\newif\ifdaiVisible
\daiVisibletrue   

\def\BibTeX{{\rm B\kern-.05em{\sc i\kern-.025em b}\kern-.08em
    T\kern-.1667em\lower.7ex\hbox{E}\kern-.125emX}}
\begin{document}

\title{Privacy-Preserving Industrial Ergonomics: mmWave-Based Automated REBA Scoring and Pose Estimation\\
}

\author{\IEEEauthorblockN{Xuhan Zhang}
\IEEEauthorblockA{\textit{Dept. of Applied AI and Robotics} \\
\textit{Aston University}\\
Birmingham, United Kingdom \\
0009-0000-9596-7633}
\and
\IEEEauthorblockN{Luis J. Manso}
\IEEEauthorblockA{\textit{Dept. of Applied AI and Robotics} \\
\textit{Aston University}\\
Birmingham, United Kingdom\\
0000-0003-2616-1120}
\and
\IEEEauthorblockN{Victor Chang}
\IEEEauthorblockA{\textit{Dept. Business Analytics and Information Systems} \\
\textit{Aston University}\\
Birmingham, United Kingdom\\
0000-0002-8012-5852}
\and
\IEEEauthorblockN{Zhuangzhuang Dai}
\IEEEauthorblockA{\textit{Dept. of Applied AI and Robotics} \\
\textit{Aston University}\\
Birmingham, United Kingdom \\
0000-0002-6098-115X}
}

\maketitle

\begin{abstract}
Work-related Musculoskeletal Disorders (WMSDs) require continuous ergonomic assessments. While Rapid Entire Body Assessment (REBA) is a gold-standard observation tool, manual monitoring is labor-intensive, and vision-based automation leads to privacy concerns. This paper proposes a novel end-to-end multi-task learning framework for privacy-preserving ergonomic assessment using millimetre-wave (mmWave) radar. A spatio-temporal backbone reconstructs 3D human skeletons, which serves as the biomechanical foundation for a subsequent regression head to generate REBA risk scores. To overcome the sparsity of radar point clouds, we utilise a multi-objective loss function incorporating biomechanical limits and temporal smoothness constraints. Furthermore, we implement an oversampling strategy to address the imbalance of high-risk postures in existing datasets. Experimental results on MMFi dataset demonstrate that our framework achieves a Categorical Accuracy of 77.78\% and real-time performance with an inference latency of 5.70 ms. Our method reaches a High-risk REBA MAE of 0.93, which significantly outperforms both direct regression and two-stage pipelines in high-risk scenarios, providing a robust solution for non-invasive industrial ergonomic assessment.
\end{abstract}


\section{Introduction}
Work-related Musculoskeletal Disorders (WMSDs) are a leading cause of occupational injuries worldwide, affecting the muscles, bones, nerves, and joints of the workforce \cite{WMSDs_risk_factors_review}. 

While ergonomists rely on observation assessment tools, such as Rapid Entire Body Assessment (REBA) \cite{REBA_origin}, to quantify postural strain, manual REBA assessment is labor-intensive and subject to observer's bias. Conversely, direct measurement methods (e.g., wearable goniometers, push/pull force sensors \cite{assess_review}) interfere workers' natural postures in industrial settings.

Automated postural assessment via computer vision has emerged as a solution, but its performance deteriorates under poor lighting conditions and occlusions, and raises significant privacy concerns. Radio Frequency (RF) sensing offers a solution to these challenges, particularly millimetre-wave (mmWave) radar. By emitting radio waves and analysing returned echoes for spatial information, mmWave radar preserves worker privacy while remaining robust to smoke, dust and varying lighting conditions. Meanwhile, mmWave operates in a high-frequency band (76-81 GHz)\cite{TI_mmwave_found}, which offers higher spatial resolution than other RF sensors. Beyond industrial scenes, it is also widely used for human recognition within Ambient Assisted Living contexts (\cite{AR_for_AAL2024},\cite{AR_Real_world2025}). 

Despite its potential, the sparsity and noise of mmWave radar point clouds challenges the angular precision for ergonomic assessment. Advanced deep learning algorithms show promise in reconstructing human poses and positions from sparse mmWave data. While two-stage pipelines, which predict skeletal coordinates and then calculate ergonomic scores, offer a modular approach, our preliminary investigation show several drawbacks:
\begin{itemize}
    \item Error Propagation: Minimal joint angle deviations can lead to significant REBA score jumps. 
    \item Architecture Complexity: Decoupled architectures require synchronisation between the pose estimator and the scoring logic, complicating real-time deployment. 
    \item Misaligned Optimisation Objective: Standard skeleton reconstructions are optimised for Euclidean distance rather than angular precision required for ergonomic risk assessment. 
\end{itemize}

We propose a privacy-preserving, sequential end-to-end multi-task framework for mmWave-based REBA scoring. With ergonomic regression conditioned on reconstructed skeletal geometry, our framework ensures that generated risk assessments are biomechanically grounded. Our contributions include:

\begin{itemize}
    \item \textbf{MmWave-based REBA Scoring Pipeline:} We present a novel framework that directly bridges raw mmWave radar data to automated REBA scoring, as a non-invasive solution for continuous ergonomic assessment. 
    \item \textbf{Sequential Multi-task Architecture:} We apply a hierarchical architecture that uses a spatio-temporal backbone to reconstruct 3D skeletal joints, which are subsequently mapped to REBA scores. The sequential dependency ensures that ergonomic predictions are interpretable and grounded in observable poses. 
    \item \textbf{Oversampling for High-risk generalisation:} To address data imbalance, particularly lack of high-risk actions in existing human activity datasets such as MMFi, we use oversampling strategy and demonstrate its benefits to model's generalisation in predicting high-risk events. 
\end{itemize}

Results show that our framework reduces high-risk ergonomic Mean Absolute Error (MAE) by 65.8\% (0.93 vs. 2.72) compared to traditional two-stage decoupled methods, and by 77.02\% (0.93 vs. 4.08) compared to direct regression.
\section{Related Work}
\subsection{Ergonomic Risk Assessment Tools}
While direct measurement tools (e.g., ergoniometers and force sensors) provide high precision, observational methods remain the industry standard due to their non-interference with natural workflows \cite{assess_review}. REBA is uniquely suited for dynamic industrial tasks as it provides a full-body kinematic analysis of Group A (trunk, neck, legs) and Group B (upper arms, lower arms, wrists) \cite{REBA_origin}. Unlike OWAS~\cite{OWAS_origin} or RULA~\cite{RULA_origin}, REBA offers higher angular sensitivity and incorporates dynamic activity modifiers, making it better suited for diverse service and industrial roles.


However, manual REBA suffers from subjectivity of human experts and high labor costs for continuously monitoring. Recent CV frameworks utilising MediaPipe or OpenPose \cite{assess_review} have automated this process via 3D joint extraction. However, they are frequently limited by worker privacy concerns, unstable lighting, and physical occlusions in industrial scenes. This motivates the need for a non-optical, privacy-preserving sensing modality like mmWave radar. 

\subsection{MmWave-based Human Pose Estimation}


MmWave radar has emerged as a robust and privacy-preserving alternative to optical sensors for human sensing. MmPose~\cite{mmPose2020} utilises CNN-based structure by projecting point clouds into 2D depth maps, losing 3D spatial fidelity. Researchers then move towards processing raw point cloud directly. PointNet \cite{pointnet} and the following PointNet++ \cite{pointnet++} learn geometric features directly from unordered and sparse point sets. To further address temporal instability and noise, RadHAR \cite{RadHAR2019} and MARS \cite{MARS2021} voxelise sparse point clouds over a sliding window  and feeds the data into a time-distributed CNN and Bi-LSTM model. HuPR \cite{HuPR2023} introduces Pose Refinement Graph Convolutional Network (GCN) to model the physical constraints of human joints. RT-Pose \cite{RT_Pose2024} uses high-dimensional 4D radar tensors to improve resolution, while mmMesh \cite{mmmesh2021} applies parametric human templates to achieve visually impressive results for full-body mesh reconstruction. However they both suffer from expensive computation and are optimised for general pose visualisation without specific angular precision required for ergonomics.

Critically, a gap exists in directly bridging mmWave radar with ergonomic assessment protocols. No existing work optimise skeletal reconstruction specifically for the biomechanical requirements of automated REBA scoring. This paper addresses this gap by proposing an end-to-end framework that optimises skeletal reconstruction and REBA risk score regression from mmWave point clouds.

\section{Methodology}

The proposed method maps a sequence of sparse mmWave point clouds $\mathbf{P} = \{P_1, \dots, P_T\}$ to continuous ergonomic risk scores $R = \{r_1, \dots, r_T\}$. We use a Hierarchical Multi-Task Learning (H-MTL) approach, to ensure that ergonomic assessments are based on physical reality:
\begin{equation}
    \hat{S}=\mathcal{F}_{pose}(P)
\end{equation} 
\begin{equation}
    \hat{R}=\mathcal{F}_{reba}(\hat{S})
\end{equation} 

\noindent
where $\mathcal{F}_{pose}$ is a spatio-temporal encoder that reconstructs the 3D skeletal pose $\hat{S}$ from point clouds, and $\mathcal{F}_{reba}$ regresses risk scores $\hat{R}$ for the sequence. 

\subsection{MmWave Radar Signal Processing}

We use a mmWave radar, operating in the 77-81 GHz band, to capture human motion. The fundamental sensing unit is the chirp, where frequency increases linearly over time. The distance $D$ to a target is derived from the beat frequency $f_b$ between the transmitted and received signals:
$$D = \frac{c \cdot f_b}{2 \cdot K}$$
where $c$ is the speed of light and $K$ is the chirp slope. A standard signal processing chain is applied to extract useful information:
\begin{itemize}
    \item 3D Fast Fourier Transform (3D-FFT): FFTs are performed to obtain range, Doppler velocity and angle of arrival respectively.
    \item Point Cloud Extraction: Cell-Averaging Constant False Alarm Rate (CA-CFAR) algorithm is applied to extract lightweight point clouds. It adaptively calculates a noise-based threshold to filter out noise and only retain major reflections:
    $$P_t = \text{CFAR}(\text{FFT}_{3D}(\text{Signal}_{IF}))$$
\end{itemize}

Unlike dense radar tensors, point clouds provide a lightweight representation that inherently masks facial identifiers, ensuring worker privacy while maintaining the necessary geometric information for Pose Estimation. Therefore, point clouds have been a very popular modality for RF-based human sensing (\cite{mmPose2020},\cite{mmmesh2021},\cite{an2022fast},\cite{sengupta2020nlp}, etc.). The sparsity significantly reduces the computational cost, and serves as a balancer between data utility for skeleton reconstruction and user privacy requirements.

\subsection{Ergonomic Ground Truth and Dataset Optimisation}\label{method:dataset}

\begin{figure}
    \centering
    \includegraphics[width=\linewidth]{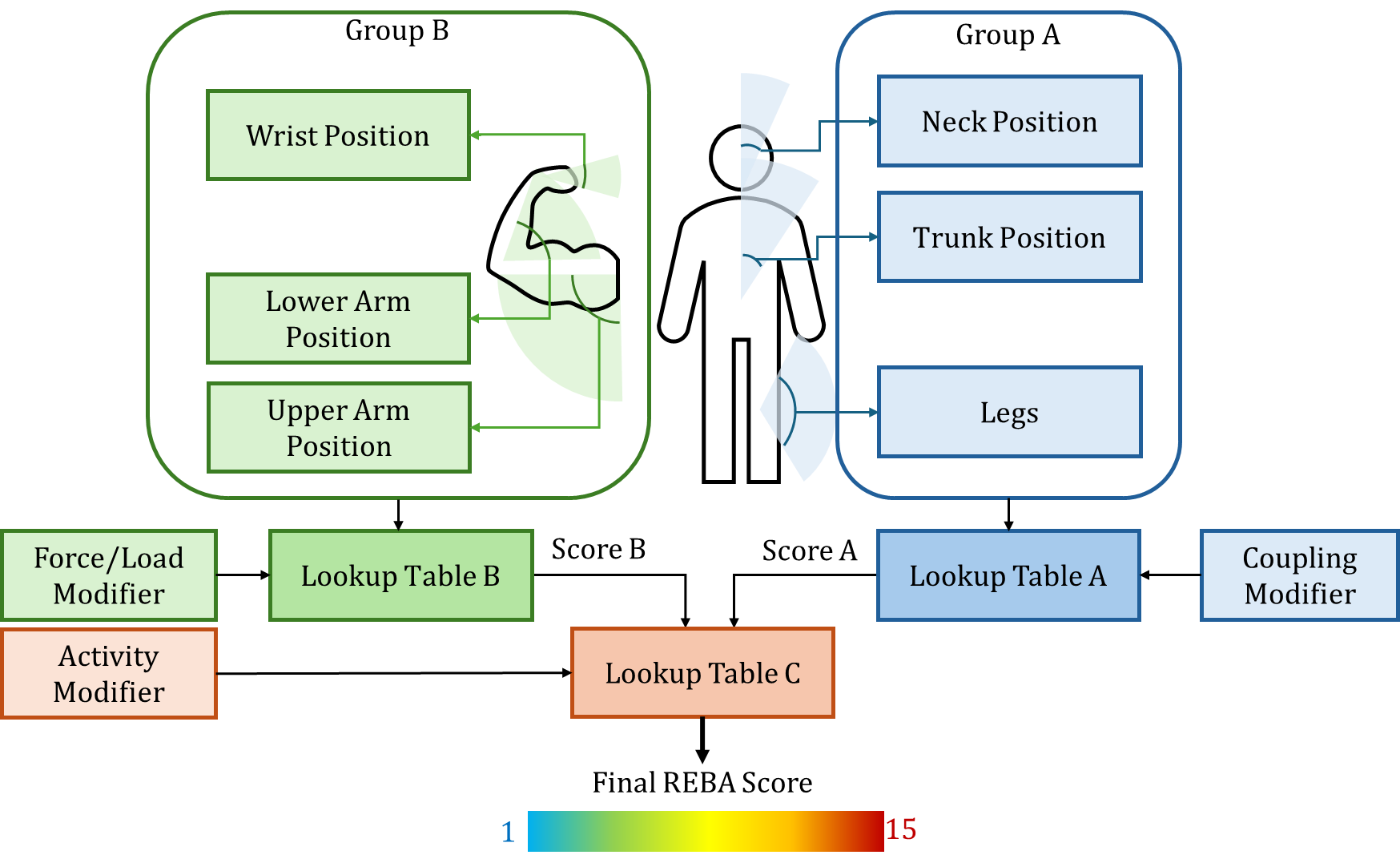}
    \caption{REBA Assessment Protocol. Postural risk is calculated by scoring two groups: Group A (blue) and Group B (green). The circular sectors specify angular scores used for cross-referencing in standarised Lookup Tables, which are further combined with external modifiers. The resulting score (1 to 15) categorises the posture into several risk levels, from negligible (1) to very high (11+).}
    \label{fig:reba}
\end{figure}

\textbf{REBA Ground Truth Generation:} To supervise the multi-task network, ground truth REBA scores $R_{GT}$ are generated by applying standardised REBA protocols to synchronised LiDAR skeletal data $S_{GT}$. As detailed in Fig. \ref{fig:reba}, we convert 3D skeletal joints into unit vectors $\vec{v}_{ij}$ to calculate relative body angle $\theta$ as:
$$\theta = \arccos\left(\frac{\vec{v}_1 \cdot \vec{v}_2}{\|\vec{v}_1\| \|\vec{v}_2\|}\right)$$
These angles are mapped to categorical postural scores (Score A, Score B) which are then cross-referenced using standardised \textbf{REBA lookup tables}. This deterministic process incorporates external modifiers (load/force and coupling) to produce the final ground truth scores $R_{GT} \in [1, 15]$ for each frame.

\textbf{Dataset Imbalance and Oversampling:} We use the MMFi dataset \cite{MMFi}, which provides synchronised mmWave radar data and high-precision skeletal annotation (LiDAR data) across 27 daily activities.

\begin{table}[htbp]
\caption{REBA Risk Distribution across Representative MMFi Actions}
\begin{center}
\renewcommand{\arraystretch}{1.4}
\begin{tabularx}{\columnwidth}{ZYYYYY}
\hline
\textbf{Action ID} & \textbf{Total Frames} & \textbf{Negligible\%} & \textbf{Low\% (2-3) } & \textbf{Med\% (4-7)   } & \textbf{High\% (8+)}\\
\hline
A19 & 11880 & 0.0 & 8.5 & 85.9 & 5.6 \\
A27 & 11880 & 0.0 & 13.8 & 84.4 & 1.7 \\
A20 & 11880 & 2.9 & 53.5 & 43.4 & 0.1\\
A24 & 11880 & 0.0 & 26.8 & 73.2 & 0.0 \\
A11 & 11880 & 52.5 & 47.5 & 0.0 & 0.0\\
\hline
\end{tabularx}
\label{tab:risk-dist}
\end{center}
\end{table}

Analysis reveals a significant distribution challenge: high-risk postures ($R \geq 8$), important to industrial safety, are significantly outnumbered by low-risk actions (Table. \ref{tab:risk-dist}). To prevent the model from being biased towards low-risk, an oversampling strategy is applied during training:
\begin{itemize}
    \item Tier 1: Actions with relatively more high-risk frames (e.g., A19, A27) are oversampled by a factor of 4.
    \item Tier 2: Actions with frequent medium-to-high risk transitions (e.g., A20, A24) are oversampled by a factor of 2.
\end{itemize}


\subsection{End-to-end Network Architecture}

\begin{figure*}[t]
    \centering
    \includegraphics[width=\textwidth]{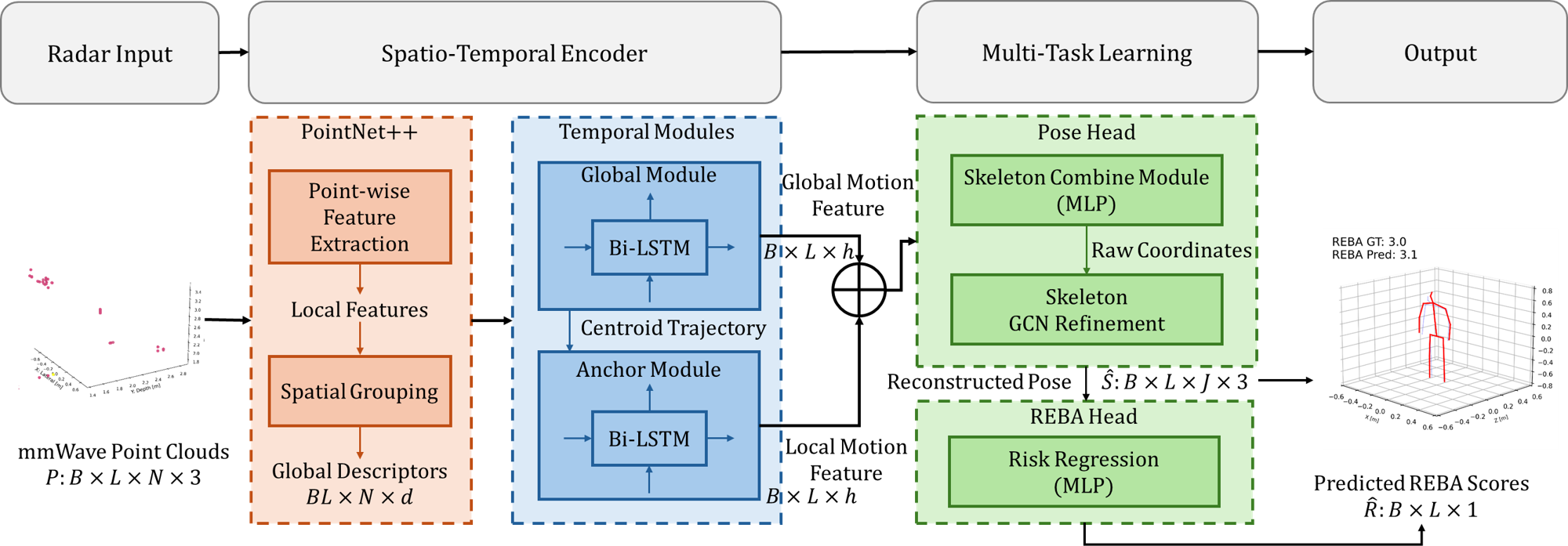}
    \caption{Overview of the Spatio-Temporal End-to-End Framework. The pipeline consists of: (1) Spatio-temporal Encoder: \textbf{PointNet++} (coral) extracts spatial features from point clouds, followed by \textbf{Temporal Modules} (blue) with dual Bi-LSTM branches that model temporal features; (2) Multi-task Learning: \textbf{Pose Head} (green) generates 3D skeletons and refines with GCN, then \textbf{REBA Head} regresses ergonomic risk scores. Notation:$B=32$ (batch size), $L=40$ (sequence length), $N=128$ (point cloud size), $d=28$ (feature frame dimension), $h=64$ (hidden dimension), $J=17$ (number of joints). }
    \label{fig:framework}
\end{figure*}

As illustrated in Fig.\ref{fig:framework}, the network is designed to map unordered point clouds to structured REBA assessments through a sequential hierarchy: 

\textbf{Spatio-Feature Encoder:} We use a PointNet++ backbone for hierarchical geometric extraction \cite{pointnet++}. Two Set Abstraction (SA) layers perform iterative sampling and grouping, converting the input point cloud to a global spatial descriptor of dimension $d=28$. This bottleneck dimension is selected to condense the high-dimensional point cloud into a compact latent representation that captures the essential spatial relationships of 17 skeletal joints while maintaining the latency for real-time processing. 

\textbf{Temporal Module:} Ergonomic risk is inherently dynamic, which means snapshots often fail to capture the activity modifier required by REBA protocols. To model the dynamics, we use a dual-path Bi-LSTM network. It is chosen to learn the past and future frame contexts within a window of $L=40$ to reduce the \textit{temporal jitter} commonly seen in sparse radar data. It contains: 
\begin{itemize}
    \item \textbf{Global Module:} Includes a Softmax Attention layer to focus on key frames inside the temporal window, allowing the model to prioritise \textit{peak} postural moments.
    \item \textbf{Anchor Module: } Sets local spatial anchors to maintain positional stability.
\end{itemize}
By treating the REBA assessment as a Markovian process where current ergonomic risk is conditioned on the preceding motion sequence, our model provides fluid and high-fidelity risk profile that prevents the \textit{score-jumping} commonly seen in snapshot assessments. The temporal module outputs the global and local motion features, and then concatenates into a spatio-temporal feature vector for the following skeletal reconstruction. 

\textbf{Multi-Task Prediction Heads:} The prediction tasks are in two tiered hierarchy where the REBA assessment is conditioned on the skeletal output:
\begin{itemize}
    \item \textbf{Pose Head:} It maps features to $J=17$ joint coordinates. To improve the physical and biological consistency as a human skeleton, we integrate a Graph Convolutional Network (GCN) that refines the joint coordinates with neigbouring joint information.
    \item \textbf{REBA Head:} It takes the refined skeletal joints as direct input, and learns to approximate skeletal geometry to the complex, non-linear mapping of REBA lookup tables. This sequential dependency ensures a predicted high-risk score is corresponding to an observable posture in the reconstructed pose. 
\end{itemize}

\subsection{Multi-Objective Loss Function}\label{loss_function}

The framework is trained using a multi-objective weighted loss function. The loss is calculated by comparing the outputs of both the Pose Head ($\hat{S}$) and REBA Head ($\hat{R}$) against the LiDAR-generated ground truth ($S_{GT},R_{GT}$ as specified in Section \ref{method:dataset}). This allows the ergonomic labels to backpropagate through the network and refine skeletal reconstruction. The total loss is defined as:
$$L_{total} = \lambda_{pos} L_{pos} + \lambda_{bone} L_{bone} + \lambda_{angle} L_{angle}$$
$$+ \lambda_{vel} L_{vel} + \lambda_{acc} L_{acc} + \lambda_{REBA} L_{REBA}$$
where each component is defined as:
\begin{itemize}
    \item $L_{pos}$: Skeletal position loss -- Mean Squared Error of 3D joint coordinates ($\hat{S}$ vs. $S_{GT}$).
    \item $L_{bone},L_{angle}$: Biomechanical constraints -- $L_{bone}$ ensures consistency in bone length across frames, while $L_{angle}$ penalises deviations in joint angles $\theta$ (Section \ref{method:dataset}).
    \item $L_{vel},L_{acc}$: Temporal loss -- minimising the first and second-order derivatives of joint positions across frames.
    \item $L_{REBA}$: Ergonomic supervision -- L1 loss between predicted and ground truth REBA scores ($\hat{R}$ vs. $R_{GT}$). 
\end{itemize}
During training, the inputs for these losses are from the final stages of the pipeline in Fig. \ref{fig:framework}. The skeletal loss components supervise the output of Pose Head, while the ergonomic loss supervises the REBA Head output. 
The code and post-processed dataset will be released upon paper acceptance.

\section{Experiments}

\subsection{Experimental Setup}
\textbf{Dataset and Subject Independence:} The pipeline is evaluated on the MMFi dataset \cite{MMFi}. To ensure subject-independent generalisability, the dataset is partitioned by subject: 864 sequences (32 subjects) are used for training, and 216 sequences (8 subjects) are reserved for testing. This ensures that model is evaluated on unseen physiological structure and movement patterns. The oversampling strategy in Section \ref{method:dataset} is applied to the training set, duplicating 320 samples to reach a more balanced training set with 1184 sequences.

\textbf{Implementation Details:} The model is implemented in PyTorch and trained using the Adam optimiser with the learning rate $10^{-4}$ for 500 epochs. We utilise a batch size of 32 and a temporal window of $T=40$ frames. To balance the multi-objective loss function in Section \ref{loss_function}, the weights are empirically tuned to balance the disparate magnitudes of the loss components, ensuring that no single component dominates the loss. Specifically, we set as follows: $\lambda_{pos} = 1.0$, $\lambda_{bone} = 0.2$, $\lambda_{angle} = 0.5$, $\lambda_{vel} = 0.2$,$\lambda_{acc} = 0.2$, and $\lambda_{REBA} = 0.5$.

\textbf{Evaluation Metrics:} To evaluate the skeletal reconstruction, we utilise 4 key metrics:
\begin{itemize}
    \item \textbf{MPJPE:} Mean Per Joint Position Error (cm) to measure absolute 3D coordinate accuracy.
    \item \textbf{Categorical Accuracy:} The percentage of frames correctly classified into REBA risk tiers (Negligible, Low, Medium, High).
    \item \textbf{REBA MAE:} Mean Absolute Error between predicted and ground truth REBA scores.
    \item \textbf{RoM Ratio:} The ratio of predicted vs. ground truth Range of Motion (RoM) variance. The RoM value closer to 1.0 indicates better preservation of dynamic movement.
\end{itemize}

\subsection{Comparison with Alternative Benchmarks}

To justify the chosen sequential end-to-end architecture, we compare the Proposed pipeline against two identical-setup baselines in Table. \ref{tab:benchmarks}:
\begin{itemize}
    \item \textbf{mmWave Direct Regression:} Maps point clouds directly to REBA scores frame-by-frame, omitting skeletal constraints.  
    \item \textbf{mmWave 2-Stage decoupled approach:} Uses the same backbone to predict 3D skeletal joints, which are then passed to an offline REBA calculation module.
\end{itemize}

\begin{table}[htbp]
\caption{Comparison with Alternative Architectures}
\begin{center}
\renewcommand{\arraystretch}{1.4}
\begin{tabularx}{\columnwidth}{p{3cm}YYY}
\hline
\textbf{Metrics}& \textbf{Direct Regression} & \textbf{mmWave 2-stage}  & \textbf{Proposed} \\
\hline
MPJPE (cm) $\downarrow$ &-- & \textbf{7.31}  &8.71    \\
REBA MAE $\downarrow$  &0.88 & 0.78 &\textbf{0.66}    \\
Categorical Accuracy $\uparrow$ &55.82\% &68.10\%  &\textbf{77.78\%} \\
High-risk MPJPE $\downarrow$&-- &16.72  &\textbf{9.04}  \\
High-risk REBA MAE $\downarrow$ &4.08 &2.72  &\textbf{0.93}  \\
Inference time (ms) $\downarrow$&\textbf{0.91} &10.40 &5.70 \\
\hline
\end{tabularx}
\label{tab:benchmarks}
\end{center}
\end{table}

As shown in Table \ref{tab:benchmarks}, the Proposed model achieves superior performance in both overall and high-risk scenarios. It reaches a Categorical Accuracy of 77.78\% which is significantly higher among them. For high-risk postures, while 2-stage pipeline suffers from error propagation leading to REBA failures (MAE of 2.72), our Proposed method reaches a High-risk REBA MAE of 0.93. Although the 2-stage pipeline shows better global MPJPE, it largely degrades in high-risk scenarios (7.31 cm increasing to 16.72 cm), whereas the Proposed model remains robust (8.71 cm to 9.04 cm). Finally, while the 5.70 ms inference frame-rate is twice as fast as the 2-stage pipeline, online deployment has a 40-frame buffer delay.

\subsection{Ablation Study and Impact of Oversampling}

We evaluate the contribution of several loss components and the oversampling strategy. The Proposed configuration includes the multi-objective loss with weights ($\lambda_{pos}=1.0$, $\lambda_{bone}=0.2$, $\lambda_{angle}=0.5$, $\lambda_{vel}=0.2$, $\lambda_{acc}=0.2$, $\lambda_{REBA}=0.5$ and tiered oversampling.

\begin{table}[htbp]
\caption{Performance with/without Oversampling}
\begin{center}
\renewcommand{\arraystretch}{1.4}
\begin{tabularx}{\columnwidth}{ZYYYY}
\hline
\textbf{Method} & \textbf{Global MPJPE (cm)} $\downarrow$ & \textbf{High-risk MPJPE (cm)} $\downarrow$ & \textbf{High-risk REBA MAE} $\downarrow$ & \textbf{Categorical Accuracy} $\uparrow$ \\
\hline
Imbalanced & \textbf{8.55} & 12.35 & 1.29 & 77.78\% \\
Proposed & 8.71 & \textbf{9.04} & \textbf{0.93} & 77.78\%  \\
\hline
\end{tabularx}
\label{tab:oversample}
\end{center}
\end{table}

\begin{figure}[t]
    \centering
    \begin{subfigure}[b]{0.24\textwidth}
        \centering
        \includegraphics[width=\textwidth]{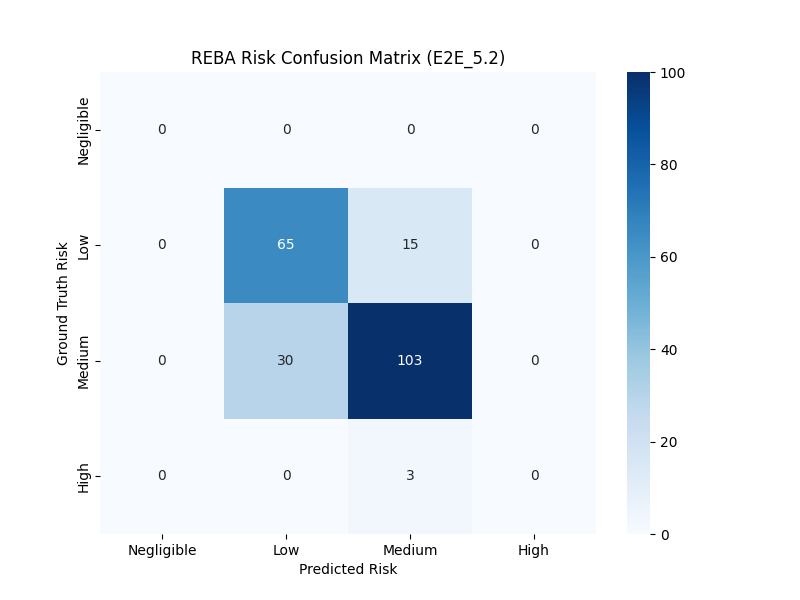}
        \caption{Without Oversampling}
        \label{fig:cm_baseline}
    \end{subfigure}
    \hfill
    \begin{subfigure}[b]{0.24\textwidth}
        \centering
        \includegraphics[width=\textwidth]{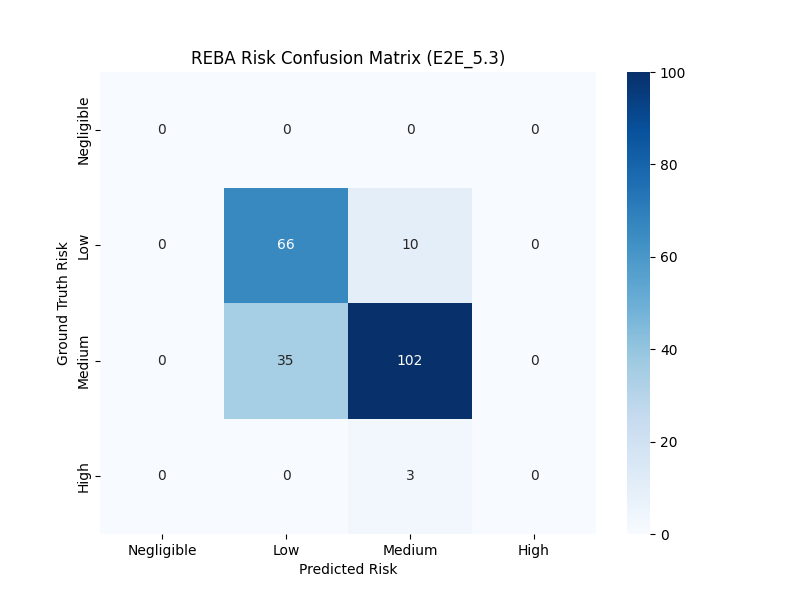}
        \caption{Proposed with Oversampling}
        \label{fig:cm_best}
    \end{subfigure}
    \caption{REBA Risk Confusion Matrices. Both models show a similar classification distribution due to the sparsity of point cloud data. }
    \label{fig:cm_comparison}
\end{figure}

\textbf{Role of Oversampling:} As shown in Table \ref{tab:oversample}, the oversampling strategy benefits the model's performance on high-risk postures. While the Imbalanced model without oversampling achieves a slightly lower global MPJPE and identical categorical accuracy, it suffers from degradation in high-risk scenarios. Conversely, Proposed method reaches a High-risk REBA MAE of 0.93. This indicates that although these two models share similar classification results (as shown in Fig. \ref{fig:cm_comparison}) due to the sparsity of point clouds, oversampling forces the model to learn more from rare and high-risk scenes.

\textbf{Ablation Study on Loss Components:} We conduct ablation studies to validate the necessity of the multi-objective loss function in Section \ref{loss_function}. The performance of each configuration is summarised in Table. \ref{tab:ablation}.

\begin{table}[htbp]
\caption{Ablation Study of Loss Function Components}
\begin{center}
\renewcommand{\arraystretch}{1.4}
\begin{tabularx}{\columnwidth}{p{2.8cm}YYYY}
\hline
\textbf{Metrics} & \textbf{Proposed} & \textbf{W/O Temporal Loss} & \textbf{W/O Angle Loss} & \textbf{High Risk Weight} \\
\hline
MPJPE (cm) $\downarrow$ & \textbf{8.71} & 8.96 & 9.89 & 9.31  \\
REBA MAE $\downarrow$ & \textbf{0.66} & 0.70 & 0.69 & 0.69  \\
RoM Ratio $\uparrow$ & 0.69 &\textbf{0.75} &0.49 &0.64 \\
Categorical Accuracy $\uparrow$ & \textbf{77.78\%} &66.20\% &76.39\% &71.30\% \\
High-risk MPJPE $\downarrow$ &\textbf{9.04} &10.05 &13.03 &12.36 \\
High-risk REBA MAE $\downarrow$ &\textbf{0.93} &1.13 &1.24 &1.22 \\
\hline
\end{tabularx}
\label{tab:ablation}
\end{center}
\end{table}

\textbf{Temporal Loss:} Removing temporal-related components ($\lambda_{vel}, \lambda_{acc} = 0$) results in lower categorical accuracy and increased high-risk error. While it achieves the highest RoM ratio, the resulting motion lacks semantic consistency. 

\textbf{Angle Loss:} Removing the angle loss ($\lambda_{angle} = 0$) degrades reconstruction significantly, with the highest global MPJPE and the lowest RoM ratio. The model fails to respect realistic skeletal limits, especially in high-risk postures.

\textbf{Risk Robustness vs. Geometry:} Increasing the REBA loss weight ($\lambda_{REBA} = 1.0$) is intended to force the model to focus on extreme postures. However, it results in worse performance in high-risk scenarios. This indicates that over-weighting the risk score makes the model prioritise the final risk level at the cost of skeletal geometry that generates it. The Proposed configuration ($\lambda_{REBA} = 0.5$) provides the most balanced performance.

\subsection{Qualitative Visualisation}

\begin{figure}[t]
    \centering
    \begin{subfigure}[b]{0.5\textwidth}
        \centering
        \includegraphics[width=\textwidth]{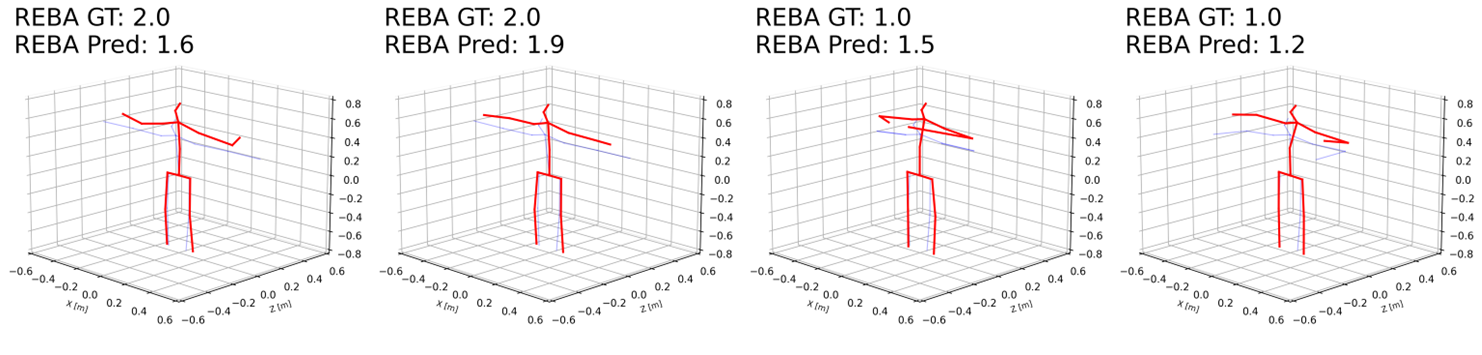}
        \caption{A02}
        \label{fig:a02}
    \end{subfigure}
    \hfill
    \begin{subfigure}[b]{0.5\textwidth}
        \centering
        \includegraphics[width=\textwidth]{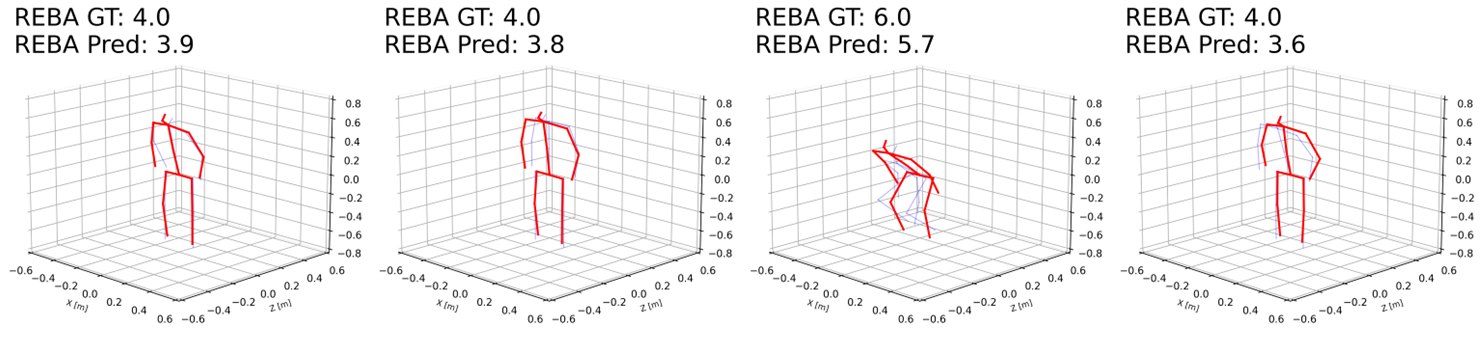}
        \caption{A19}
        \label{fig:a19}
    \end{subfigure}
        \begin{subfigure}[b]{0.5\textwidth}
        \centering
        \includegraphics[width=\textwidth]{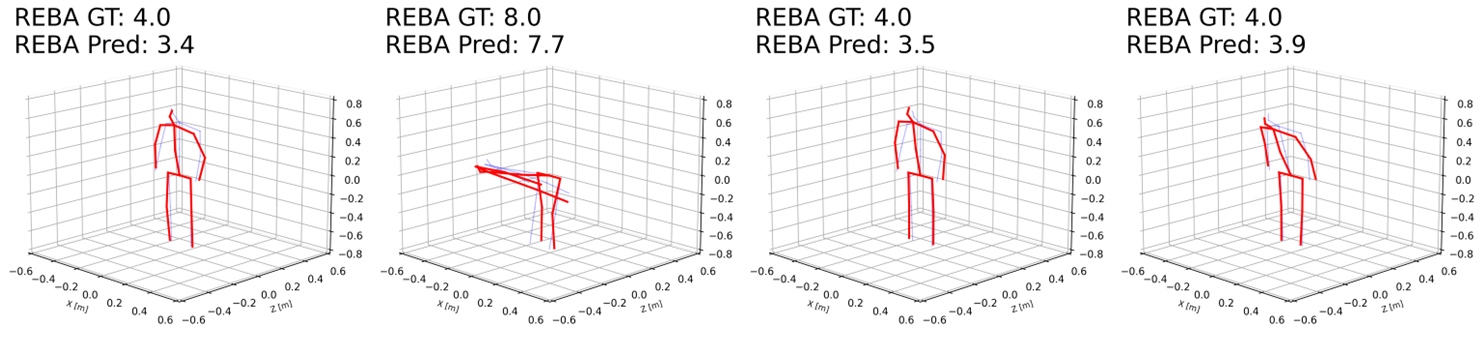}
        \caption{A27}
        \label{fig:a27}
    \end{subfigure}
    \caption{Visualised results across different actions. (Red: Predicted; Blue: Ground Truth) (a) A02 (Extending Arms): precise upper-limb tracking. (b) A19 (Squatting): robustness to self-occlusion and whole-body dynamics. (c) A27 (Bending): accurate capture of high-risk postures.}
    \label{fig:result_frames}
\end{figure}

The predicted 3D skeletons and synchronised REBA scoring results are presented in Fig. \ref{fig:result_frames}. (sampled at frames 5, 15, 25 and 35). These sequences demonstrate the model's performance on various actions. Specifically, A19 (squatting) shows resilience to body self-occlusion, while A27 (bending) shows tracking during high-risk ergonomic transitions.

\section{Conclusion}
This work presents a novel end-to-end framework for privacy-preserving ergonomic assessment using mmWave radar. By conditioning REBA regression on 3D skeletal reconstruction via a multi-objective loss, we ensure risk scores are biomechanically grounded. Our oversampling strategy improves robustness in high-risk scenarios. The pipeline demonstrates clear feasibility for real-time industrial ergonomic assessments. 

To ensure reproducibility, our full implementation will be made publicly available on GitHub.

\textbf{Limitations and Future Work:}
\begin{itemize}
    \item \textbf{Data and Action Diversity:} Evaluation is limited to MMFi. Future work will incorporate datasets like mmRadPose \cite{mmRadPose2025} and industrial actions.
    \item \textbf{Ergonomic Modifiers:} Current public mmWave datasets lack non-kinematic variables like load and coupling, forcing our model to rely on kinematics. Future work involves collecting custom datasets with those factors.
    \item \textbf{Hardware Resolution:} Sparse point clouds limit joint precision; higher-resolution radar may further reduce localisation errors.
\end{itemize}
\bibliographystyle{IEEEtran}
\bibliography{references}

\end{document}